\documentclass[letterpaper, 10 pt, journal, twoside]{IEEEtran}
 \IEEEoverridecommandlockouts %
\usepackage[T1]{fontenc}
\usepackage[utf8]{inputenc}

\usepackage{amsmath} %
\usepackage{amssymb} %
\usepackage{bbm}

\usepackage{graphicx}

\usepackage{booktabs}
\usepackage{amsmath}
\usepackage{siunitx}
\usepackage{multirow}

\usepackage[sort,compress]{cite}

\usepackage{bm}
\usepackage{url}
\usepackage{bbold}
\usepackage{hyperref}
\usepackage{cleveref}
\usepackage{microtype}

\DeclareMathOperator*{\argmax}{arg\,max}

\usepackage{ifthen}

\newif\ifarxiv
\arxivtrue
\usepackage{eso-pic}
\usepackage{url}
\AddToShipoutPicture*{%
     \AtTextUpperLeft{%
         \put(-3.5,10){
           \begin{minipage}{\textwidth}
              \scriptsize
              \MakeUppercase{Preprint version of an IEEE Robotics and Automation Letters article; final version available at} \url{https://doi.org/10.1109/LRA.2025.3641119}
           \end{minipage}}%
     }%
}

\title{Ensemble-Based Event Camera Place Recognition\\Under Varying Illumination}

\ifarxiv
\author{Therese Joseph\hskip5em Tobias Fischer\hskip5em Michael Milford%
\thanks{Manuscript received: August 24, 2025; Revised  November 4, 2025; November 18, 2025. This paper was recommended for publication by Editor Sven Behnke upon evaluation of the Associate Editor and Reviewers' comments.}
\thanks{The authors are with the QUT Centre for Robotics, School of Electrical Engineering and Robotics, Queensland University of Technology, Brisbane, QLD 4000, Australia. Email: {\tt\footnotesize t2.joseph@hdr.qut.edu.au}}%
\thanks{This research was partially supported by the QUT Centre for Robotics and funding from ARC Laureate Fellowship FL210100156 to MM and ARC DECRA Fellowship DE240100149 to TF.}%
\thanks{Digital Object Identifier (DOI): see top of this page.}
}
\else
\author{}
\fi

\begin{document}

\maketitle

\bstctlcite{BSTcontrol}

\begin{abstract}

Compared to conventional cameras, event cameras provide a high dynamic range and low latency, offering greater robustness to rapid motion and challenging lighting conditions. Although the potential of event cameras for visual place recognition (VPR) has been established, developing robust VPR frameworks under severe illumination changes remains an open research problem. Here, we introduce an ensemble-based approach to event camera place recognition that combines sequence-matched results from multiple event-to-frame reconstructions, VPR feature extractors, and temporal resolutions. Unlike previous event-based ensemble methods, which only utilise temporal resolution, our broader fusion strategy delivers significantly improved robustness under varied lighting conditions (e.g., afternoon, sunset, night), achieving up to 77\% relative improvement in Recall@1 across day-night transitions. We evaluate our approach on two long-term driving datasets (with 8 km per traverse) without metric subsampling, thereby preserving natural variations in speed and stop duration that influence event density. We also conduct a comprehensive analysis of key design choices, including binning strategies, reconstruction methods, and feature extractors, to identify the most critical components for robust performance. 
Additionally, we propose a modification to the standard sequence matching framework that enhances performance at longer sequence lengths. To facilitate future research, we release our codebase and benchmarking framework\footnote{\url{https://github.com/theresejoseph/ensemble_event_vpr_bench}}.

\end{abstract}


\begin{figure*}[t]
\vspace{6pt}
  \centering
  \includegraphics[width=\textwidth]{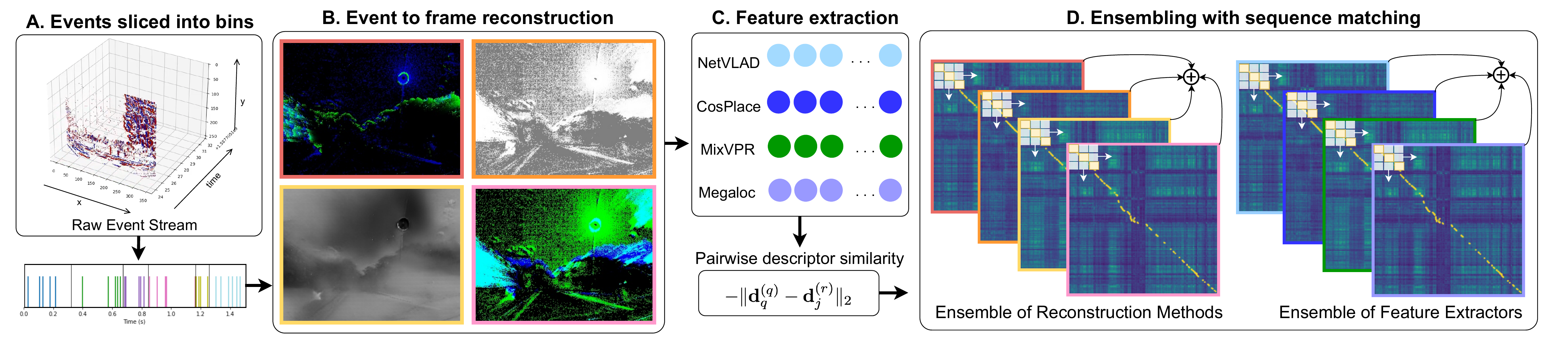}
  \vspace{-1cm}
  \caption[Pipeline for Ensemble based Event Camera Place Recognition under Varied Illumination]{
\textbf{Overview of our ensemble-based event camera place recognition pipeline.}
A) Each bin is reconstructed into a 2D frame using one of several methods: event count (with/without polarity), time surface~\cite{lagorce_hots_2017}, or the learned E2VID model~\cite{rebecq_high_2019}.
(B) These reconstructed frames are processed by an ensemble of visual place recognition (VPR) feature extractors (NetVLAD~\cite{arandjelovic_netvlad_2018}, CosPlace~\cite{berton_rethinking_2022}, MixVPR~\cite{ali-bey_mixvpr_2023}, MegaLoc~\cite{berton_megaloc_2025}) to generate global descriptors.
(C) Pairwise descriptor similarities are computed, followed by sequence matching on each resulting similarity matrix.
(D) Finally, sequence scores are aggregated across the multiple reconstruction methods, feature extractors, and temporal resolutions to enhance recognition performance.
}
  \label{fig:pipeline}
  \vspace{-0.5cm}
\end{figure*}

\section{Introduction}
\IEEEPARstart{E}{vent} cameras are a class of vision sensors that record changes in brightness asynchronously and independently for each pixel, producing a stream of events that encode the time, pixel location, and polarity of intensity changes. Unlike conventional cameras, they offer microsecond temporal resolution, high dynamic range, and low latency, making them particularly suitable for high-speed, low-power applications and operation in challenging lighting conditions \cite{ falanga_dynamic_2020,hines_compact_2025,gallego_event-based_2022}.

These characteristics have generated growing interest in applying event cameras to autonomous navigation tasks~\cite{lee_eventvlad_2021,fischer_event-based_2020,lee_ev-reconnet_2023,fischer_how_2022,kong_event-vpr_2022,hines_compact_2025}, including visual place recognition (VPR), a core component of long-term localisation and map-based re-localisation. Traditional VPR systems rely on frame-based imagery and aim to extract features invariant to appearance changes, dynamic elements, viewpoint shifts, and lighting variation~\cite{masone_survey_2021,lowry_visual_2016,zhang_visual_2021,schubert_visual_2024,garg_where_2021}. However, they often falter under extreme illumination changes (e.g., day-night)\cite{lengyel_zero-shot_2021}, where event cameras can still provide a useful signal. Despite their promise, event-based VPR remains underexplored, especially in conditions involving severe illumination variation. Key open research questions include: 1) how to best represent and reconstruct event data for downstream place recognition, and 2) whether combining complementary representations or VPR models can improve robustness under challenging conditions. 

We present an ensemble-based approach to event camera VPR that enhances robustness through late score fusion. Specifically, we propose three ensemble strategies: (i) aggregating across different event-to-frame reconstruction methods, both classical (e.g., event count, time surface~\cite{lagorce_hots_2017}) and learned (E2VID~\cite{rebecq_high_2019}); (ii) aggregating across multiple VPR feature extractors, including NetVLAD~\cite{arandjelovic_netvlad_2018}, CosPlace~\cite{berton_rethinking_2022}, MixVPR~\cite{ali-bey_mixvpr_2023}, and MegaLoc~\cite{berton_megaloc_2025}, and (iii) aggregating all configurations with temporal resolutions, event-to-frame reconstruction methods and VPR feature extractors.

Unlike prior event-based VPR ensembling that fuses predictions only across temporal resolutions~\cite{fischer_event-based_2020}, our strategy evaluates ensembles of reconstruction methods, feature extractors and temporal resolutions, yielding improved robustness under challenging appearance changes, including illumination shifts across the day. This broader fusion combines complementary cues from different representations, leading to greater robustness across diverse visual and motion conditions. We also use complete traverses without metric subsampling in our evaluations, which preserves natural variations in speed that are challenging in event-based settings where event density is motion-dependent. Finally, we apply a modified sequence matching strategy to each ensemble member, leveraging temporal consistency for improved performance.

\noindent In this work, we contribute:
\begin{enumerate}
\item An ensemble-based event VPR framework that aggregates predictions across reconstruction methods, feature extractors and temporal resolutions, achieving up to 77\% relative gain in Recall@1 under day–night transitions.

\item A comprehensive analysis of event-based VPR design choices, including binning strategies (time- vs.~count-based), polarity inclusion, reconstruction methods and feature extractor performance.

\item A modification to sequence matching from \cite{milford_seqslam_2012} and \cite{garg_seqmatchnet_2022} with dynamic
history length and z-score normalisation for improved performance at longer sequence lengths. 

\item Extensive evaluation on two challenging long-term driving datasets, \textit{BrisbaneEventVPR}~\cite{fischer_event-based_2020} and \textit{NSAVP}~\cite{carmichael_dataset_2025} under varied lighting conditions. 
\end{enumerate}

\section{Related Works}
\label{sec:relatedworks}
In this section, we review related works, focusing on three core areas that inform our contributions:  conventional VPR methods using frame-based imagery (\Cref{rel_work:VPR}), event-based VPR approaches, which adapt place recognition pipelines to event camera data (\Cref{rel_work:EVPR}), and ensemble techniques in VPR (\Cref{rel_work:ensemb_VPR}), which aim to improve robustness and performance through fusion of models or representations.

\subsection{Place Recognition}
\label{rel_work:VPR}

 Place recognition matches a query to a reference database via feature extraction and similarity search, enabling global re-localisation and loop closure~\cite{masone_survey_2021,lowry_visual_2016,zhang_visual_2021,schubert_visual_2024,garg_where_2021,tsintotas_revisiting_2022}. Visual place recognition (VPR) uses vision-based sensors such as conventional frame or event cameras. Beyond vision, LiDAR-based methods learn or extract geometry-aware point cloud descriptors~\cite{zhang_lidar-based_2024}, radar-based methods explore robustness to weather and illumination~\cite{venon_millimeter_2022}, and multimodal methods combine camera frames with LiDAR or events~\cite{melekhin_mssplace_2025,komorowski_minkloc_2021,hou_fe-fusion-vpr_2023}.

 Modern VPR relies on deep networks: NetVLAD (VGG-16 + VLAD)~\cite{arandjelovic_netvlad_2018}, CosPlace (ResNet-50 with disjoint classes)~\cite{berton_rethinking_2022}, MixVPR (MLP mixing on flattened features)~\cite{ali-bey_mixvpr_2023}, and MegaLoc (DINOv2-Base + SALAD)~\cite{berton_megaloc_2025}. In our work, we adopt these methods for event-based place recognition, evaluating both their performance and the benefits of ensembling.

\subsection{Event-Based VPR}
\label{rel_work:EVPR}
Event cameras output a continuous stream of sparse, asynchronous events rather than conventional image frames. Compared to conventional frame-based cameras, they provide microsecond-level temporal resolution, very low latency, high dynamic range, and low power consumption, which makes them well-suited to mobile sensing under fast motion and challenging illumination conditions such as night driving or headlight glare. These sensing properties are directly beneficial for visual place recognition, where robustness to motion blur, abrupt lighting changes, and low-light scenes is essential for reliable matching along long vehicle traverses~\cite{wang_event_2025, chakravarthi_recent_2025}.

To leverage standard vision pipelines, particularly those based on deep learning, these events are typically aggregated into frame-like representations or voxel tensors. An early study on event-based visual SLAM~\cite{milford_towards_2015} demonstrated the feasibility of place recognition over a 2.7km route using SeqSLAM~\cite{milford_seqslam_2012} applied to polarity-removed event count reconstructions. Fischer et al.\cite{fischer_event-based_2020} later introduced an ensemble-based approach that fused multiple temporal windows, reconstructing event frames with E2VID\cite{rebecq_high_2019} and extracting features using NetVLAD. In follow-up work~\cite{fischer_how_2022}, they proposed a lightweight method using a sparse subset of varying pixels as descriptors and sum of absolute differences (SAD) for matching, achieving competitive performance with minimal compute cost.

Subsequent works have adopted learning-based paradigms. Lee et al.\cite{lee_eventvlad_2021} introduced a semi-supervised network that reconstructs denoised edge images using GRUs and convolutional modules. Ev-ReconNet\cite{lee_ev-reconnet_2023} is a CNN-based autoencoder for edge reconstruction, which was later converted into a spiking neural network (SNN) for neuromorphic deployment. Both methods use NetVLAD for retrieval. Kong et al.\cite{kong_event-vpr_2022} proposed a weakly supervised, end-to-end VPR framework that converts event bins into event spike tensor (EST) voxel grids\cite{gehrig_end--end_2019}, extracts features with a ResNet-34 backbone, and applies a VLAD layer with triplet ranking loss. While their dataset suite included night traverses, they only reported explicit day–night evaluation on the synthetic Oxford RobotCar and Carla simulator dataset.

Although Ensemble-Event-VPR~\cite{fischer_event-based_2020} and Event-VPR~\cite{kong_event-vpr_2022} conducted ablations on windowing strategies, reconstruction methods, network backbones and loss functions, several aspects remain underexplored. In particular, prior work offers limited comparison of time surface reconstructions, event binning methods, and the performance of various SoTA VPR feature extractors when applied to event data, especially under day–night transitions. Our work addresses these gaps through an evaluation of these design decisions, demonstrating their significant impact on VPR performance under varying illumination conditions. Beyond analysing these factors individually, we show that combining them through score-level fusion further enhances robustness, as complementary representations contribute distinct but reinforcing cues under appearance change.

Other efforts explored hardware optimisations and deployment. One study applied a closed-loop bias controller for brightness-adaptive VPR, using SAD matching on event count reconstructions\cite{nair_enhancing_2024}. Most recently, Hines et al.~\cite{hines_compact_2025} demonstrated event-based place recognition using a spike-based encoder with event count frames accumulated over a second, deployed on ultra-low-power neuromorphic hardware.

\subsection{Ensemble Methods}
\label{rel_work:ensemb_VPR}

Ensembles improve accuracy when base models are both accurate and diverse. If individual models have uncorrelated errors and perform better than random guessing, the probability that a majority vote is incorrect becomes lower than the individual error rate~\cite{hansen_neural_1990,dietterich_ensemble_2000}.

In VPR, ensembles enhance robustness at different stages of the pipeline. Hierarchical Multi-Process Fusion applies late re-ranking across pipelines~\cite{hausler_hierarchical_2020}, while Patch-NetVLAD aggregates multi-scale patch features for viewpoint changes~\cite{hausler_patch-netvlad_2021}. In neuromorphic settings, ensembles of spiking networks learn non-overlapping regions and combine spike responses at inference~\cite{hussaini_ensembles_2023}. For event-based VPR, ensembling across time resolutions has shown gains~\cite{fischer_event-based_2020}. This method identified that fusion based on the mean of similarity or difference matrices yielded the highest performance across product, median, min, max, trimmed mean and weighted strategies. 

We extend these ideas by ensembling across reconstruction methods, feature extractors, and temporal resolutions, by using late score fusion to combine the varied similarity responses of these heterogeneous configurations.  This extension enables complementary modalities to reinforce each other, improving matching consistency across changes in lighting, texture, and motion dynamics.

\section{Methodology}

Our event-based VPR pipeline, illustrated in Fig.~\ref{fig:pipeline}, consists of five main stages. Section~\ref{subsec:event_slice} details how asynchronous event data \((x, y, t, p)\) is segmented into discrete bins. Section~\ref{subsec:e2f} describes the reconstruction of each event bin into a 2D frame using one of several methods, including learned methods and methods with and without polarity. Section~\ref{subsec:feat_matching} outlines how the reconstructed frames are processed by SoTA place recognition (VPR) models to extract global descriptors and compute pairwise descriptor similarities. Section~\ref{subsec:seq_match_method} introduces a modified sequence matching algorithm with dynamic history length and z-score normalisation to improve robustness for longer sequences. Section~\ref{subsec:ensem_method} presents our general ensembling strategy, which aggregates similarity scores across reconstruction methods and feature extractors.

\subsection{Event Slicing}
\label{subsec:event_slice}
Event cameras record a continuous stream of events where each event \( e_j = (x_j, y_j, t_j, p_j) \) contains the pixel location \((x_j, y_j)\), the timestamp \(t_j\), and the polarity \(p_j \in \{-1, +1\}\), representing a positive or negative brightness change.

Following standard practice in event-based vision \cite{fischer_event-based_2020, milford_towards_2015, nair_enhancing_2024}, we evaluate two binning strategies: count-based binning and time-based binning. In the count-based binning strategy, each bin contains a fixed number of events, \( N \). Specifically, the \( i \)-th bin \( \mathbf{B}_i \) is defined as:
\begin{equation}
\mathbf{B}^{N}_i = \{ e_j \mid j \in [iN, (i+1)N) \}.
\label{eq:count_bin}
 \end{equation}
In the time-based binning strategy, each bin spans a fixed duration temporal window \( \Delta t \), starting from a reference time \( t_0 \). The \( i \)-th bin is given by:
\begin{equation}
\mathbf{B}^{t}_i = \{ e_j \mid t_j \in [t_0 + i\Delta t,\ t_0 + (i+1)\Delta t) \}.
\label{eq:time_bin}
\end{equation}

\subsection{Event to Frame Reconstruction}
\label{subsec:e2f}
Using the binned event data \( \mathbf{B}_i = \{ e_j \} \), we construct frame-based representations using several established reconstruction methods \cite{rebecq_high_2019, lagorce_hots_2017}.

The first method is a learned reconstruction model, E2VID~\cite{rebecq_high_2019}, which produces an image estimate from a sequence of events using an encoder-decoder architecture with spatiotemporal fusion. We denote the reconstructed frame from bin $\mathbf{B}_i$ as $\mathbf{I}^{\text{E2VID}}_i = \mathcal{R}_{\text{E2VID}}(\mathbf{B}_i)$, where $\mathcal{R}_{\text{E2VID}}$ is the pretrained E2VID model applied to the events in bin $\mathbf{B}_i$.

For classical, non-learned reconstructions, we implement the event count reconstruction. In the polarity-aware variant, we maintain two separate channels, one for each polarity:
\begin{equation}
\mathbf{I}^{\text{count}}_i(x, y, c) = \sum_{e_j \in \mathbf{B}_i} \mathbb{1}(x_j = x \wedge y_j = y \wedge p_j = c),
\label{eq:event_count}
\end{equation}
where \( c \in \{-1, +1\} \) denotes the polarity channel, and \( \mathbb{1}(\cdot) \) is the indicator function. In the polarity-agnostic variant, the channels are summed into a single grayscale count image.

We also construct polarity-aware time surfaces~\cite{lagorce_hots_2017} with exponential decay. For each pixel and polarity, we store the timestamp of the most recent event and compute
\begin{equation}
\mathbf{I}^{\text{TS}}_i(x, y, c) = \exp\left( -\frac{t_{\text{ref}} - t(x, y, c)}{\lambda} \right),
\label{eq:time_surface_decay}
\end{equation}
where \( t(x, y, c) \) is the timestamp of the most recent event for this pixel location and polarity, \( \lambda \) is a decay constant, and \( t_{\text{ref}} = t_{\text{end}} + \lambda \) is the adjusted frame reference time, where \( t_{\text{end}} \) is the timestamp of the last event in bin \( \mathbf{B}_i \). The offset in \(t_{\text{ref}}\) ensures that even the most recent events are partially decayed, resulting in a smoother, more expressive time surface.

For each valid bin (i.e., containing at least one event), events are passed to the selected reconstruction method, which produces either a grayscale or multi-channel image. All reconstructions are computed at the native sensor resolution. For E2VID, hot pixel suppression is applied due to its sensitivity to structured noise, which can cause artifacts by interpreting hot pixels as meaningful features. Classical methods are more robust to such noise, as their simple aggregation tends to smooth over spurious events.

\begin{figure}[t]
  \vspace*{3pt}
  \centering
  \includegraphics[width=\columnwidth]{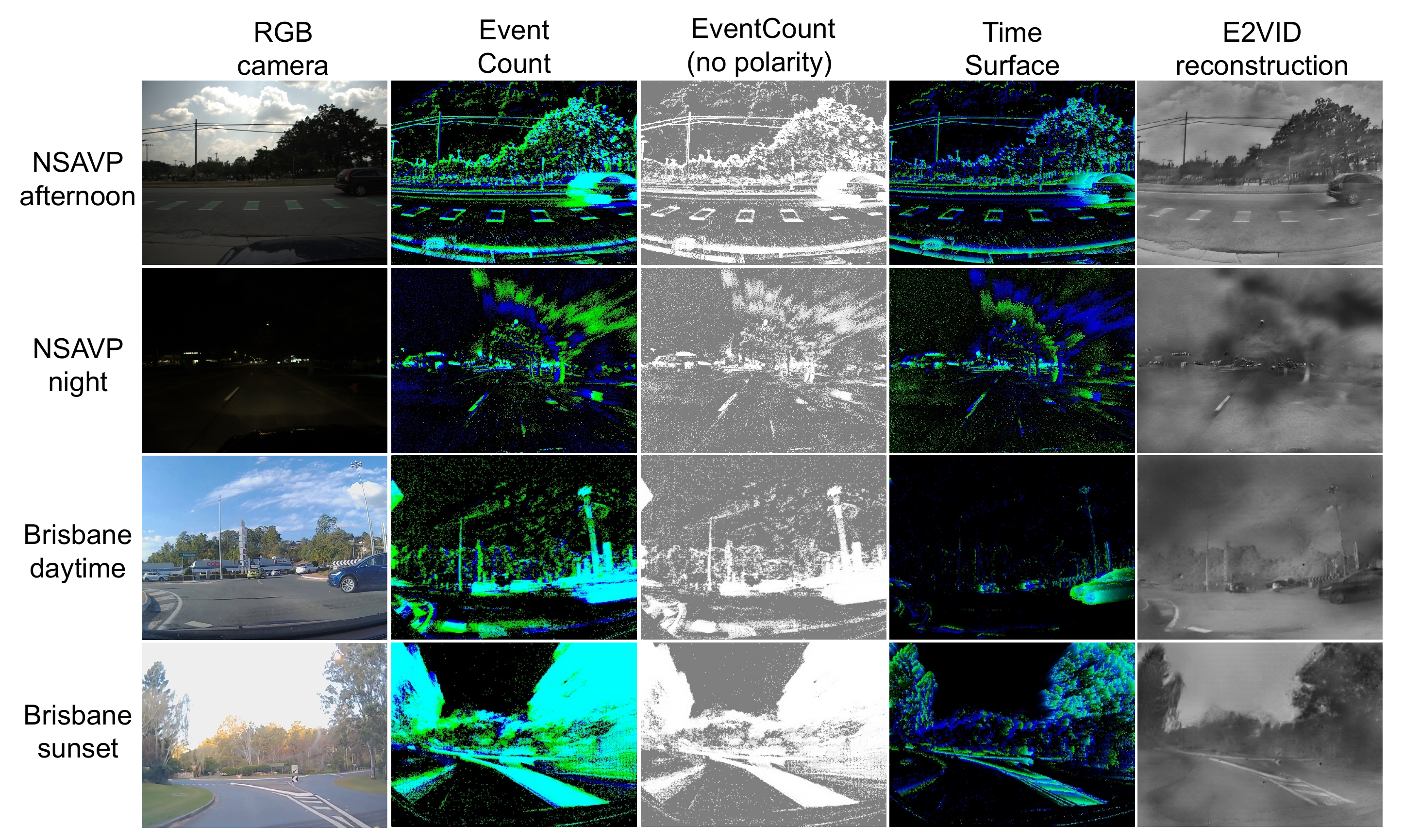}
  \vspace*{-0.4cm}
\caption[Illustration of Event Stream to Frame Reconstructions at Various Times of Day ]{Reconstructed event frames under varying illumination conditions. Column 1 shows the corresponding standard camera image. Columns 2–5 show event-based reconstructions: (2) Two-channel Event Count (polarity-separated), (3) Single-channel Event Count (polarity-combined), (4) Two-channel Time Surface (polarity-separated), and (5) Single-channel E2VID reconstruction (learned method). Rows 1–2 show afternoon and night conditions from a Gen4 Prophesee sensor in the NSAVP dataset~\cite{carmichael_dataset_2025}, while rows 3–4 show daytime and sunset conditions from a DAVIS 346 sensor in the Brisbane Event dataset~\cite{fischer_event-based_2020}.}
\vspace{-0.5cm}
  \label{fig:recon_examples}
\end{figure}

All classical reconstructed frames are normalised to produce image-like inputs by applying a per-pixel hyperbolic tangent to compress the dynamic range, followed by min--max scaling to the 8-bit grayscale range \([0, 255]\). For polarity-aware reconstructions, events are split into separate positive and negative arrays. Each is normalised using the same \(\tanh\) and scaling procedure, then assigned to separate colour channels (e.g., green for positive, blue for negative), with the third channel set to zero. This results in a three-channel image compatible with standard learned feature extractors. An example of these reconstructions is shown in Figure~\ref{fig:recon_examples}, illustrating the information captured by each method.

\subsection{Feature Extraction and Descriptor Matching}
\label{subsec:feat_matching}
We use the reconstructed frames to generate feature descriptors using models originally trained for conventional frame-based VPR. We evaluate several feature extractors, including the classical model NetVLAD~\cite{arandjelovic_netvlad_2018},  a large-scale training approach CosPlace~\cite{berton_rethinking_2022}, a lightweight architecture MixVPR~\cite{ali-bey_mixvpr_2023},  and the current SoTA model, MegaLoc~\cite{berton_megaloc_2025} (see~\Cref{rel_work:VPR}).

Let $\mathbf{I}^{Q}_q$ and $\mathbf{I}^{R}_j$ denote the $q$-th query frame and the $j$-th reference frame, respectively. A descriptor extractor $f(\cdot)$ maps each frame to a descriptor vector in a common embedding space: $\mathbf{d}^{Q}_q = f(\mathbf{I}^{Q}_q)$, $\mathbf{d}^{R}_j = f(\mathbf{I}^{R}_j)$, $\mathbf{d} \in \mathbb{R}^D$, where $D$ is the descriptor dimension, $Q$ is the size of the query set and $R$ is the size of the reference set. We compute the similarity between each query and reference descriptor using the negative L2 distance, and define the similarity matrix \( \mathbf{S} \in \mathbb{R}^{Q\times R} \) as:
\begin{equation}
\mathbf{S}[q, j] = -\lVert \mathbf{d}^{Q}_q - \mathbf{d}^{R}_j \rVert_2, \quad j_q = \argmax_j \, \mathbf{S}[q, j],
\end{equation}
where higher values of \( \mathbf{S}[q, j] \) indicate greater similarity, and \( j_q \) denotes the predicted reference match for query \( q \).

\subsection{Sequence Matching}
\label{subsec:seq_match_method}
In trajectory matching tasks with repeat traversals, the similarity matrix typically exhibits a strong diagonal, with local warping due to speed changes, stops, or route deviations. SeqSLAM
~\cite{milford_seqslam_2012} takes advantage of this structure by applying temporal consistency across the similarity matrix. \mbox{\textit{SeqMatchNet}}~\cite{garg_seqmatchnet_2022} adapted a simplified implementation of SeqSLAM by convolving a diagonal kernel across the similarity matrix, and scoring sequences based on their diagonal alignment. We use this implementation in our work. 

To improve robustness near sequence boundaries and mitigate local biases caused by visual aliasing, we extend \mbox{\textit{SeqSLAM}} by introducing a variable-length matching kernel and dual-axis normalisation. While \mbox{\textit{SeqSLAM}} computes per-row diagonal scores using a fixed-size convolutional kernel, our method adapts the effective kernel size near sequence boundaries. Let \( R \) denote the number of reference frames and \( L \) the desired sequence length. For each query index $q$, we construct a submatrix $\mathbf{C}$ of the similarity matrix $\mathbf{S} \in \mathbb{R}^{Q \times R}$ by extracting up to $L$ preceding query frames (the available history in a real time deployment setting), without exceeding the matrix boundary: $\mathbf{C}_{q} \in \mathbb{R}^{N \times R}$, where $N = \min(L, q + 1)$.

We apply z-score normalisation to \( \mathbf{C} \) across columns and then across rows, to reduce local bias in the similarity submatrix.  To compute the sequence-matched score at each reference index \( j \), we calculate the trace of the most recent \( k \times k \) (where $k = \min(N, j + 1)$) diagonal block:
\begin{equation}
\tilde{\mathbf{S}}[q,j] = \operatorname{trace}\left( \mathbf{C}_{[N - k:N,\; j - k + 1:j + 1]} \right).
\end{equation}

Using the trace is mathematically equivalent to convolving with a diagonal identity kernel, but allows for a more efficient implementation, as only the diagonal elements are computed rather than the full square kernel. This two-stage normalisation and variable-length kernel enable sequence matching to be applied even at the boundaries and improve robustness to local noise and visual aliasing. We apply this sequence matching procedure independently to each method configuration—i.e., each combination of reconstruction method and feature extractor—producing a set of sequence-matched similarity matrices that serve as inputs to our ensembling strategy.

\subsection{Ensembling Strategies}
\label{subsec:ensem_method}

Each $\tilde{\mathbf{S}}^{(m)} \in \mathbb{R}^{Q \times R}$ is the sequence-matched similarity matrix produced by the $m$-th method configuration. We treat these as individual ensemble members: $\mathcal{S} = \left\{ \tilde{\mathbf{S}}^{(1)},\ \dots,\ \tilde{\mathbf{S}}^{(M)} \right\}$.

To aggregate predictions across configurations, we perform element-wise summation of all aligned similarity matrices, followed by maximum similarity selection as per the ensembling strategy from~\cite{fischer_event-based_2020}:
\begin{equation}
\mathbf{\mathcal{E}} = \sum_{m=1}^{M} \tilde{\mathbf{S}}^{(m)}, \quad
\bar{j}_q = \argmax_j \, \mathbf{\mathcal{E}}[q, j].
\label{eq:ensemble_score_fusion}
\end{equation}

Unlike prior event-based VPR ensembling, which fuses predictions across temporal resolutions only~\cite{fischer_event-based_2020}, our strategy evaluates ensembles of reconstruction methods, ensembles of feature extractors and a combined ensemble of feature extractors, reconstruction and temporal ensembles. This broader fusion leads to significantly improved robustness under challenging appearance changes, including illumination shifts across the day.

\vspace{-0.15cm}
\section{Experimental Setup}
\label{sec:experimentalsetup}

\subsection{Datasets}

In this paper, we evaluate our method on two outdoor datasets: BrisbaneEventVPR\cite{fischer_event-based_2020} and NSAVP\cite{carmichael_dataset_2025}, recorded during real-world vehicle traversals across day and night conditions.

BrisbaneEventVPR is a widely used benchmark dataset for event-based visual place recognition, introduced in 2020 and evaluated in several works~\cite{fischer_how_2022, lee_eventvlad_2021, lee_ev-reconnet_2023, hines_compact_2025}. It consists of six traverses of an 8km route in Brisbane, Australia, recorded under various weather and illumination conditions, including daytime, night, sunrise, sunset, and morning on single and multilane roads. Data was captured using a $346\times260$ resolution iniVation DAVIS 346 event camera mounted on the vehicle’s windshield, along with GPS and 1080p RGB video from a consumer-grade camera.

Novel Sensors for Autonomous Vehicle Perception (NSAVP) is a more recent dataset collected in Michigan, USA, with a focus on evaluating modern sensor technologies in diverse and realistic environments. It comprises two distinct routes (8.3 km and 8.6 km) traversed in both directions under varying lighting conditions. Data was collected using a $1280\times720$ resolution Prophesee Gen4 HD event sensor and high-precision 200 Hz ground-truth from an Applanix POS-LV 420 system, yielding a higher-resolution, roof-mounted configuration from a different manufacturer than the DAVIS 346 used in BrisbaneEventVPR. Compared to BrisbaneEventVPR, NSAVP also introduces greater environmental diversity, including suburban neighbourhoods and dense urban scenes with multi-story buildings, and offers higher fidelity sensing for large-scale benchmarking. 

To ensure fair comparison across illumination changes, we evaluate five day-to-day and five day-to-night pairs. 
On BrisbaneEventVPR~\cite{fischer_event-based_2020}, we use \texttt{sunset1} as the reference against \texttt{sunrise}, \texttt{morning}, \texttt{daytime}, \texttt{sunset2}, and \texttt{night}. 
On NSAVP~\cite{carmichael_dataset_2025}, we adopt the notation \texttt{R0-[F/R]-[A/S/N]} (Route 0, Forward/Reverse, Afternoon/Sunset/Night) to evaluate five pairs: FA0 vs. FS0, FA0 vs. FN0, FN0 vs. FS0, RA0 vs. RN0, and RS0 vs. RN0. 
Datasets such as DDD20~\cite{hu_ddd20_2020} and NYC Events~\cite{pan_nyc-event-vpr_2024} were excluded as they lack explicit repeat traversals.

 \subsection{Implementation Details}
We adopt \textit{Recall@1} as our primary evaluation metric, following standard practice in the VPR literature. This metric measures the ratio of correct top-1 matches to the total number of query frames. 

To evaluate event binning strategies, we experiment with both count-based and time-based binning. All subsequent experiments use time-based binning, which consistently yields stronger performance. For each reconstructed frame, we linearly interpolate available GPS data to determine its ground-truth position. We extract features using the open-source \textit{VPR-Methods-Evaluation} repository, applying a 25-metre tolerance to define correct matches~\cite{berton_eigenplaces_2023}. This tolerance is smaller than previous methods \cite{fischer_event-based_2020, carmichael_dataset_2025} for the Brisbane Event dataset, which used 70m, so our reported recall results are more conservative. No feature extractor or reconstruction model is retrained; all models are used as released. For comparison to LENS~\cite{hines_compact_2025}, we use 1 second temporal windows with event count reconstruction, a ground truth tolerance of +/- 3 seconds and a sequence length of 10. 

For ensemble evaluation, we follow the protocol of~\cite{fischer_event-based_2020}, which requires temporal alignment between ensemble members. We consider time resolutions of 0.1, 0.25, 0.5, and 1.0 seconds, sampled at 1~Hz to match the coarsest resolution. Each single ensemble strategy, based on time resolution, reconstruction method, or feature extractor, is evaluated across 480 configurations ($4\times 4\times 3 \times10$). These configurations are generated by combining 4 feature extractors, 4 reconstruction methods, 4 time resolutions, 3 sequence lengths (10, 20, 30), and 10 reference-query traverse pairs. We also evaluate a configuration that aggregates across all combinations of the individual methods (a total of 64) and report on these results as the combined ensembles.

\section{Results and Discussion}
\label{sec:results}
This section presents and compares the performance of our event-based VPR system. Section~\ref{sub_sec:esem_perf} contains our evaluation of each ensemble strategy across reconstruction methods and feature extractors. Section~\ref{subsec:seq_results} investigates the effect of our modified sequence matching approach (Section~\ref{subsec:seq_match_method}) compared to the original SeqSLAM methodology. The influence of event-binning strategies on recognition accuracy and robustness is then examined in Section~\ref{subsec:event_bin}. Finally, Section~\ref{sub_sec:perf_feat_recon} compares key design choices in event-based VPR, including binning strategies, polarity, reconstruction methods, and feature extractors, critically evaluating the impact of each component on performance.

\subsection{Event VPR via Ensembling}
\begin{figure}[t]
  \centering
  \includegraphics[width=0.95\columnwidth]{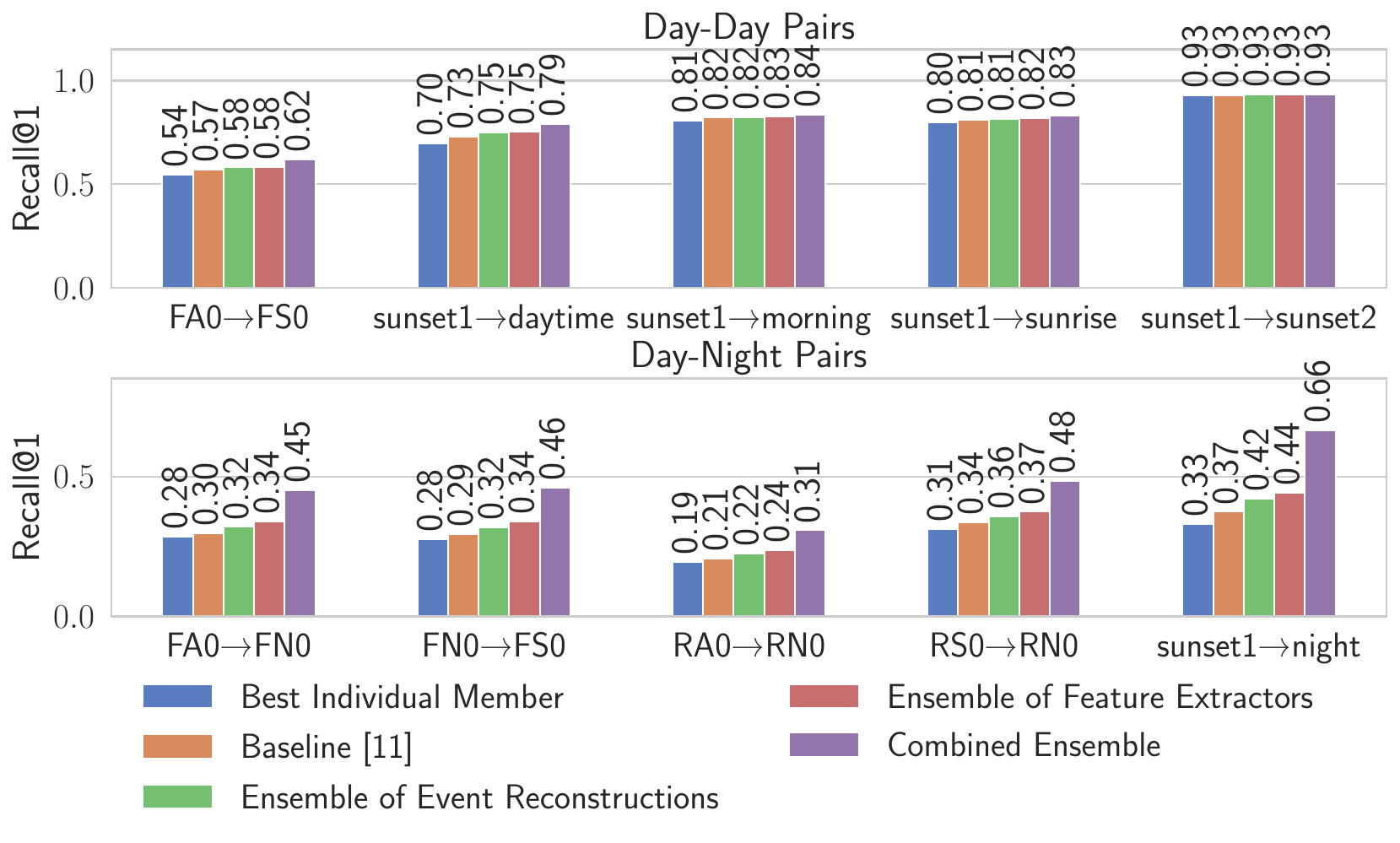}
  \vspace*{-0.3cm}
  \caption{Average Recall@1 (AR@1) for each reference–query pair across ensembling strategies and best individual method, shown separately for day and night conditions. Results are averaged over sequence lengths of 10, 20, and 30 at 1 Hz sampling. The combined ensemble aggregates predictions from varied feature extractors, event-to-frame reconstructions and temporal resolutions.}
  \vspace*{-0.5cm}
  \label{fig:ensem_perf}
\end{figure}

The effectiveness of the proposed ensemble strategies across reconstruction methods, feature extractors and the combination of feature extractors, reconstruction and temporal ensembles was evaluated against two baselines: (i) the best-performing individual ensemble member and (ii) the temporal resolution only ensemble from \cite{fischer_event-based_2020}. The results exhibit a bimodal distribution, reflecting the expected performance gap between day–day and day–night pairs due to significant domain shifts. To highlight these trends, results are grouped into day and night conditions. As shown in Figure~\ref{fig:ensem_perf}, across 10 diverse traverse combinations, the proposed ensembles consistently outperformed both baselines.

\begin{figure}[t]
  \centering
  \vspace{0.2cm}
  \includegraphics[width=0.9\columnwidth]{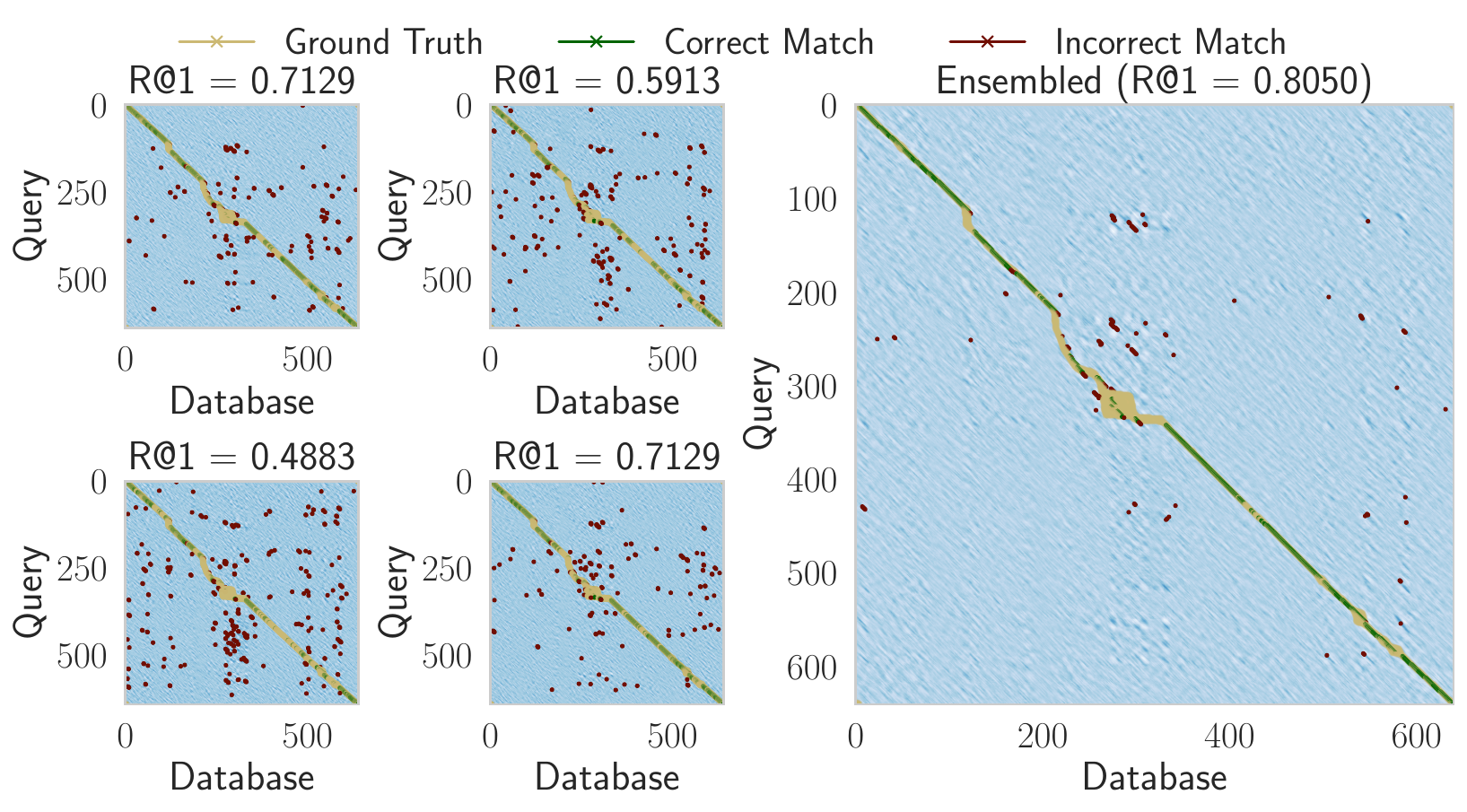}
  \caption{Similarity matrices for a sunset–daytime pair with Megaloc feature extractor and varied reconstructions. Columns 1 and 2 show results from E2VID, EventCount, EventCount (no polarity) and time surface reconstructions. Column 3 shows the ensemble of reconstruction methods. Ground-truth, correct, and incorrect matches are annotated.}
  \label{fig:esem_exam}
  \vspace*{-0.2cm}
\end{figure}

Table~\ref{tab:imprv_time_ensem} quantifies comparisons to existing methods along with the relative improvement over the temporal ensemble baseline~\cite{fischer_event-based_2020}. Under severe illumination variation on BrisbaneEventVPR (day to night), the proposed ensembling improves average recall@1 (AR@1) from 1\% for LENS~\cite{hines_compact_2025} and 37.41\% for Temporal Ensembles~\cite{fischer_event-based_2020} to 44.16\% and 66.39\% for the ensemble of feature extractors and combined ensembles. On NSAVP day–night, the feature extractor and combined ensembles improve AR@1 from 28.40\% to 32.18\% and to 42.55\%, respectively. Therefore, the combined ensembles have a relative improvement of up to 77\% in day-night transitions. Gains in daytime conditions are smaller but consistent, with up to 3.02\% relative gain on BrisbaneEventVPR and 8.63\% relative gain on NSAVP. Additionally, figure~\ref{fig:esem_exam} illustrates the fusion of predictions from multiple event-to-frame reconstructions, yielding a higher recall than the best individual method and highlights the robustness gained through aggregation.

\begin{table}[t]
\small
\setlength{\tabcolsep}{3pt}
\centering
\vspace{0.3cm}
\caption{Average Recall@1 and Gain over Temporal Ensemble~\cite{fischer_event-based_2020}.}
\label{tab:imprv_time_ensem}
\footnotesize
\begin{tabular}{lccc}
\toprule
\textbf{Condition} & \textbf{Method} & \textbf{AR@1 (\%)} & \textbf{Gain (\%)} \\
\midrule
\multirow{4}{*}{\shortstack{D--D\\(Brisbane\\Event)}}
  & LENS~\cite{hines_compact_2025} & 73.00 & -11.23 \\
  & Temporal Window~\cite{fischer_event-based_2020} & 82.24 & --- \\
  & Reconstruction & 82.98 & 0.91 \\
  & Feature Extractor & 83.16 & 1.13 \\
  & Combined Ensemble & 84.72 & 3.02 \\
\midrule
\multirow{4}{*}{\shortstack{D--N\\(Brisbane\\Event)}}
  & LENS~\cite{hines_compact_2025} & 1.00 & -97.33 \\
  & Temporal Window~\cite{fischer_event-based_2020} & 37.41 & --- \\
  & Reconstruction & 42.06 & 12.44 \\
  & Feature Extractor & 44.16 & 18.06 \\
  & Combined Ensemble & 66.39 & 77.48 \\
\midrule
\multirow{4}{*}{\shortstack{D--D\\(NSAVP)}}
  & Temporal Window~\cite{fischer_event-based_2020} & 56.88 & --- \\
  & Reconstruction & 58.09 & 2.13 \\
  & Feature Extractor & 58.31 & 2.51 \\
  & Combined Ensemble & 61.79 & 8.63 \\
\midrule
\multirow{4}{*}{\shortstack{D--N\\(NSAVP)}}
  & Temporal Window~\cite{fischer_event-based_2020} & 28.40 & --- \\
  & Reconstruction & 30.47 & 7.31 \\
  & Feature Extractor & 32.18 & 13.34 \\
  & Combined Ensemble & 42.55 & 49.83 \\
\bottomrule
\end{tabular}
\end{table}

\label{sub_sec:esem_perf}

\label{sub_sec:seq_match_compare}
\begin{table}[t]
\centering
\caption{AR@1 across sequence lengths and p-values from paired t-test}
\label{tab:seq_improve}
\begin{tabular}{cccr}
    \toprule
    \textbf{Seq Len} & \textbf{SeqSLAM} & \textbf{Ours} & \textbf{p-value} \\
    \midrule
    10 & 0.3473 & 0.3410 & 0.6874 \\
    20 & 0.4301 & \textbf{0.4613} & 0.0554 \\
    30 & 0.4593 & \textbf{0.4996} & 0.0096 \\
    \bottomrule
    \vspace{-1cm}
\end{tabular}
\end{table}

\subsection{Sequence Matching}
\label{subsec:seq_results}
Table~\ref{tab:seq_improve} compares the original SeqSLAM implementation (via SeqMatchNet) to the proposed sequence matching variant, which introduces an adaptive matching kernel and additional score normalisation. Results, averaged over 800 combinations of routes, reconstruction methods, feature extractors, and temporal resolutions, indicate improved Recall@1 at longer sequence lengths. At sequence lengths 20 and 30, the proposed method achieves Recall@1 gains of 0.0312 (p = 0.0554) and 0.0403 (p = 0.0096), respectively. Performance at length 10 is similar across both methods (p = 0.6874). The highest average Recall@1 is observed at a sequence length of 30 using the proposed approach.

\subsection{Event Binning Strategies}
\label{subsec:event_bin}
\begin{figure}[t]
\vspace{5pt}
  \centering
  \includegraphics[width=\columnwidth]{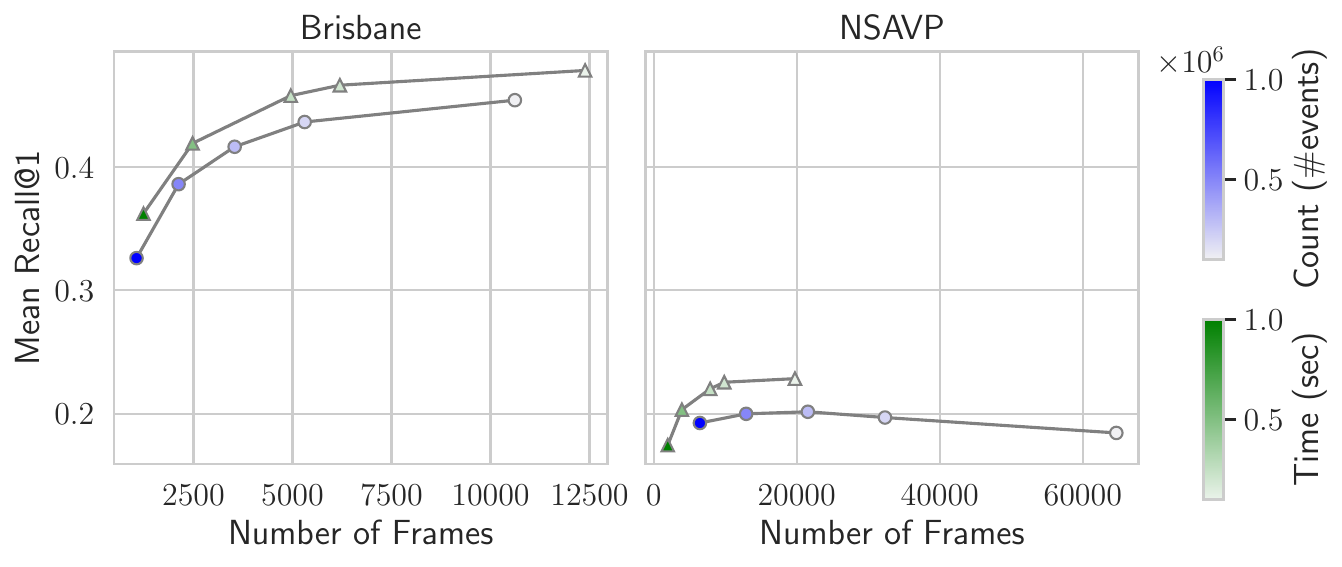}
  \caption{Average Recall@1 versus the number of reference and query frames, evaluated across two binning strategies and varying event slice resolutions. Time-based binning consistently yields the highest recall, with performance improving at higher resolutions (i.e., smaller bin sizes). The total number of frames serves as a proxy for computational cost, as both event-to-frame reconstruction and feature extraction scale linearly with frame count. Consequently, higher-resolution bins incur greater compute cost.}
  \label{fig:perf_time_count}
\end{figure}
\begin{figure}[t]
\vspace{6pt}
  \centering
  \includegraphics[width=0.9\columnwidth]{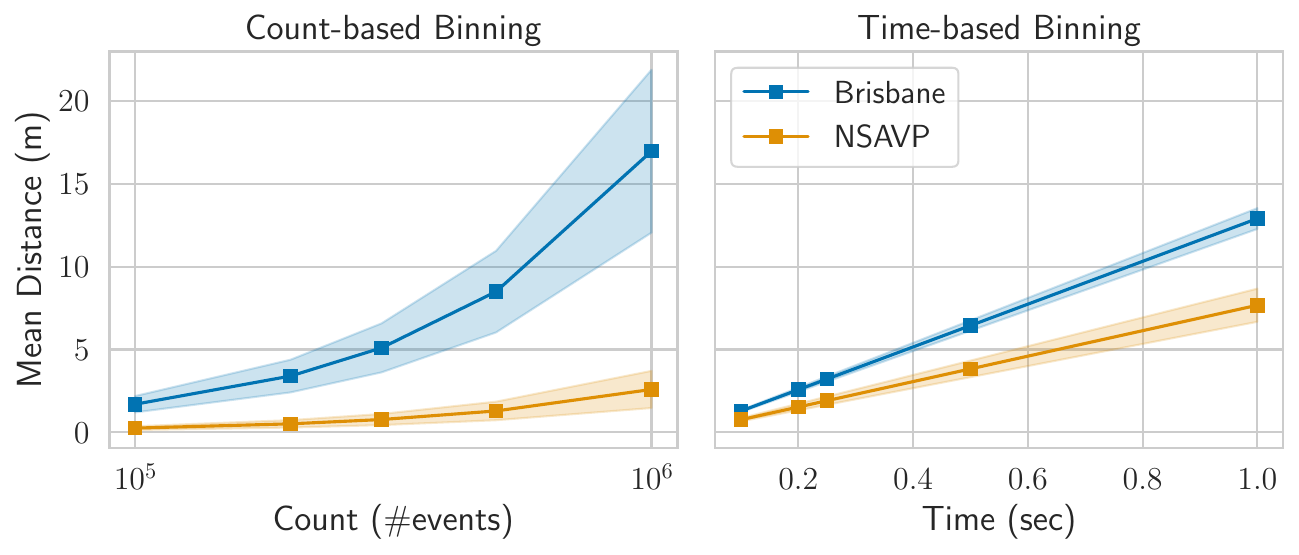}
  \caption{Distance travelled per frame averaged across repeat traverses for each binning type and resolution. Time-based binning shows the lowest variance in mean distance, indicating greater consistency for place representation. }
  \vspace{-0.5  cm}
  \label{fig:dist_time_count}
\end{figure}

Two common event binning strategies are evaluated: fixed-count (\(B^{N}\)) and fixed-duration (\(B^{t}\)). As shown in Figure~\ref{fig:perf_time_count}, fixed-duration binning consistently yields better Recall@1 in urban driving scenarios with standard VPR feature extractors. The figure also illustrates the trade-off between performance and binning resolution, where smaller bins improve accuracy but increase computational cost due to a higher number of reference and query frames. While a similar comparison was presented in~\cite{fischer_event-based_2020}, only minimal performance differences were reported, likely due to the limited temporal resolution range (0.02–0.14 seconds) used. In contrast, the broader range considered here (0.1–1.0 seconds) captures motion dynamics more effectively. To further investigate this, Figure~\ref{fig:dist_time_count} examines the distance travelled within individual bins across methods and resolutions, showing that time-based binning results in lower variance relative to the mean distance, supporting more consistent place representations across repeated traversals with varied illumination.

\subsection{Reconstructions and Feature Extractors for Event VPR}
\label{sub_sec:perf_feat_recon}
\begin{figure}[t]
\vspace{8pt}
\centering
\includegraphics[width=1\columnwidth]{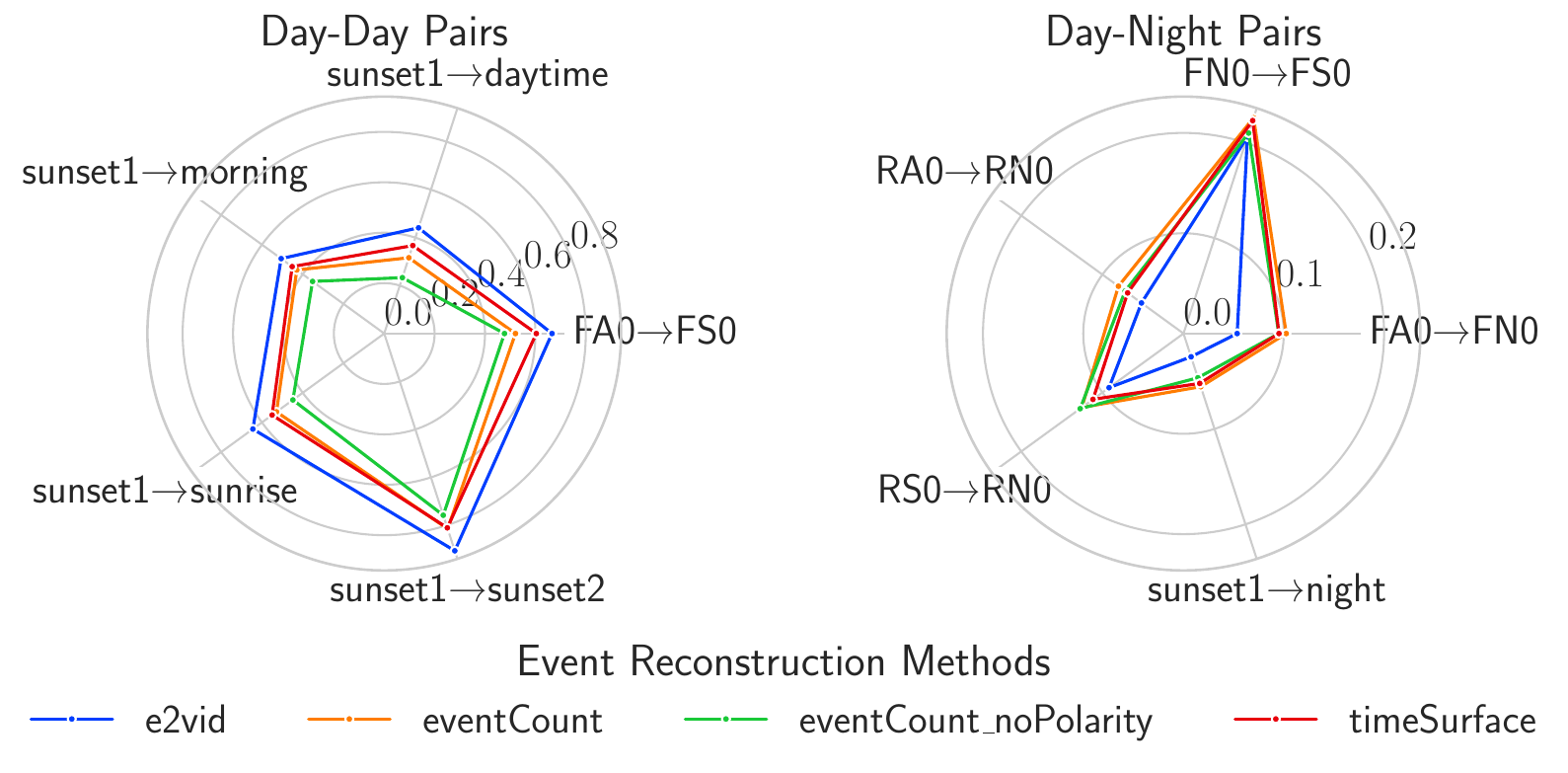}
\caption{Recall@1 across individual reconstruction methods, separated by day and night conditions, evaluated without sequence matching. These results show that the learned method E2VID performs best during the day but is detrimental at night. Encoding polarity in a separate channel also improves performance in daytime conditions for classical reconstructions.}
\label{fig:perf_recon_methods}
\end{figure}

\begin{figure}[t]
\centering
\includegraphics[width=1\columnwidth]{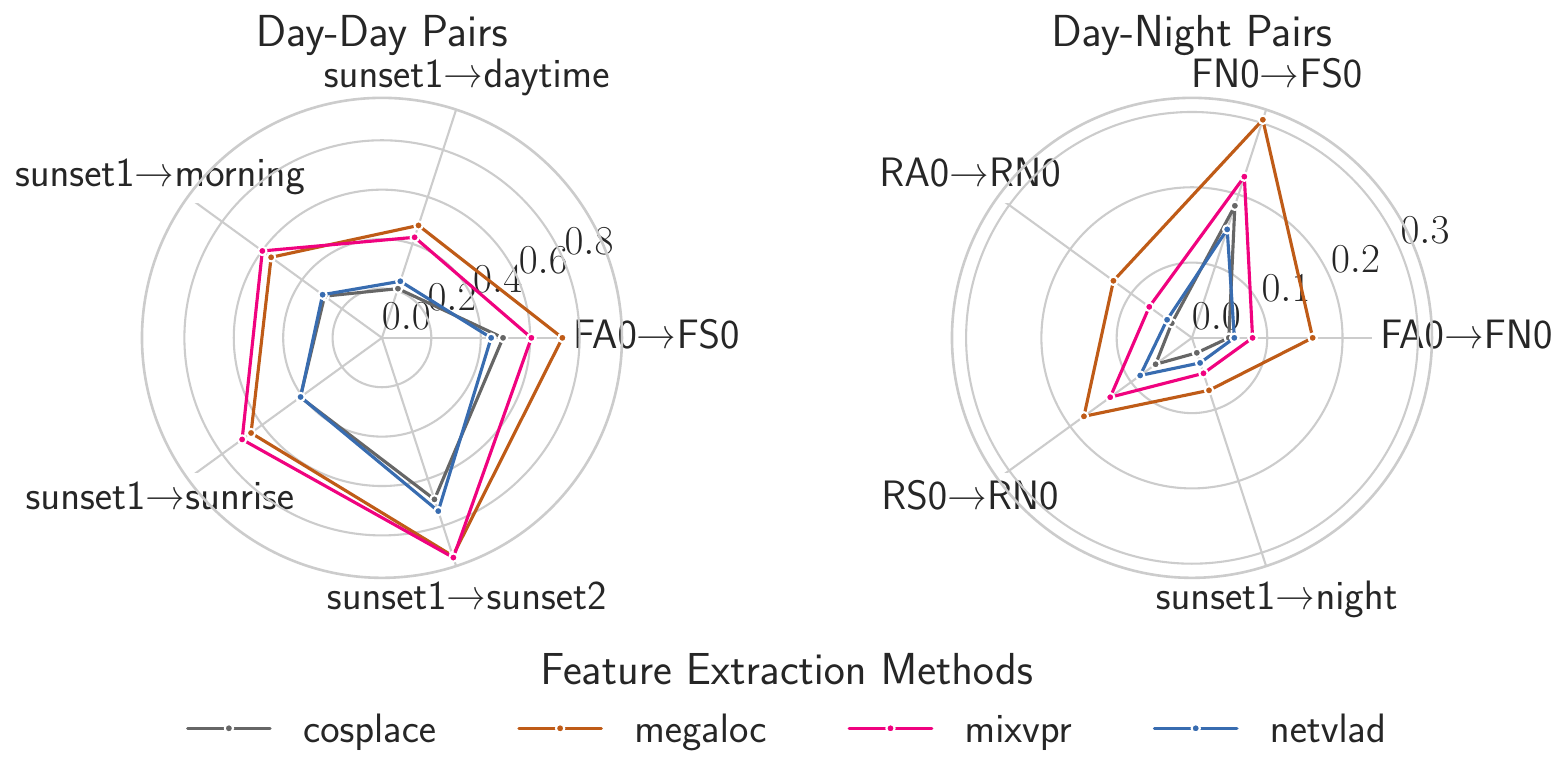}
\caption{Recall@1 across individual feature extraction methods, separated day and night conditions, evaluated using single-frame VPR without sequence matching. Event-based VPR shows similar performance trends to conventional frame-based methods, with Megaloc achieving the highest overall recall.}
\label{fig:perf_vpr_methods}
\vspace{-0.5cm}
\end{figure}

Figure~\ref{fig:perf_recon_methods} presents Recall@1 performance across individual event-to-frame reconstruction methods, averaged over five temporal resolutions.

Under daytime conditions, the learned reconstruction method E2VID achieves the highest performance, with a mean Recall@1 of 63.3\%, outperforming all classical reconstructions by 7–18\%. This is consistent with its supervised training on well-lit scenes. In contrast, under nighttime conditions, E2VID records the lowest performance (Recall@1 of 8.6\%), approximately 3–8\% lower than classical methods, which demonstrate greater robustness in low-light environments.

Two variants of event count reconstruction are also evaluated: with and without polarity. Including polarity improves daytime performance by 6.8\% Recall@1 (52.2\% vs. 45.4\%), while the difference at night is minimal (11.8\% vs. 11.0\%).
Figure~\ref{fig:perf_vpr_methods} shows VPR performance across different feature extractors. The observed trends are consistent with those reported in standard camera-based VPR benchmarks. Earlier methods, such as CosPlace and NetVLAD, achieve mean Recall@1 scores of 24–25\%, while more recent approaches like MixVPR and MegaLoc reach 38–42\%. MegaLoc, the current state-of-the-art, performs particularly well at night, achieving 16.9\% Recall@1—at least 5\% higher than all other evaluated methods.

\subsection{Computational Tradeoff}
Table~\ref{tab:runtime} reports the end-to-end runtime for a single query and AR@1 for different ensemble configurations, illustrating the accuracy versus compute tradeoff. The temporal-window~\cite{fischer_event-based_2020} and reconstruction ensembles improve AR@1 by around 6–8\% over the single model; while the feature extraction ensemble is most suitable for real-time application running at 13Hz with ~10\% increase in AR@1. The combined ensemble achieves the highest AR@1 but with substantially higher latency, making it most suitable for sparse global relocalisation or offline evaluation. A complementary correlation analysis of similarity matrices across reconstruction methods and feature extractors (included in our public code repository) shows that some representations, such as event count and event count without polarity, are strongly correlated, suggesting that future work could prioritise more complementary combinations to achieve better accuracy–runtime trade-offs.

\label{sub_sec:runtime across ensembles}
\begin{table}[t]
\vspace{0.2cm}
\centering
\caption{Runtime vs Performance Analysis}
\begin{tabular}{lcc}
\toprule
\textbf{Ensemble Type} & \textbf{Runtime (ms)} & \textbf{AR@1 (\%)} \\
\midrule
None & 47.63 & 47.07 \\
Temporal Windows~\cite{fischer_event-based_2020} & 190.5 & 52.95 \\
Event Reconstructions & 190.5 & 54.61 \\
VPR Feature Extractors & 78.38 & 56.64 \\
Combined Ensemble & 1254.0 & 64.37 \\
\bottomrule
\end{tabular}
\label{tab:runtime}
\vspace{-0.6cm}
\end{table}

\section{Conclusions and Future Work}
\label{sec:conclusion}

Despite growing interest in event-based place recognition, the full capabilities of event cameras remain underutilised. With the emergence of new long-term driving datasets, we present an ensemble-based method that fuses multiple VPR feature extractors, event reconstruction and temporal resolutions, consistently outperforming prior baselines. Beyond performance gains, our work addresses a gap in the event VPR literature by evaluating key design choices. We show that time-based binning outperforms count-based alternatives, and that the learned E2VID reconstruction, while effective during the day, performs poorly at night. Furthermore, we find that VPR performance trends across feature extractors mirror those observed in conventional frame-based settings.

While our study includes a diverse set of feature extractors, reconstructions, and temporal strategies, further gains may be achieved by learning event-specific feature extractors and developing reconstructions tailored for VPR in night conditions. Notably, none of the deep learning-based feature extractors used were trained on event data, suggesting a promising direction for developing event-native architectures. 
Future work could also explore real-world deployment of our method, expanding the utility of event cameras in challenging environments.

\bstctlcite{BSTcontrol}
{\small
\bibliographystyle{IEEEtran}
\bibliography{ReferencesLocal_abbrev}}

\begin{thebibliography}{10}
\providecommand{\url}[1]{#1}
\csname url@samestyle\endcsname
\providecommand{\newblock}{\relax}
\providecommand{\bibinfo}[2]{#2}
\providecommand{\BIBentrySTDinterwordspacing}{\spaceskip=0pt\relax}
\providecommand{\BIBentryALTinterwordstretchfactor}{4}
\providecommand{\BIBentryALTinterwordspacing}{\spaceskip=\fontdimen2\font plus
\BIBentryALTinterwordstretchfactor\fontdimen3\font minus \fontdimen4\font\relax}
\providecommand{\BIBforeignlanguage}[2]{{%
\expandafter\ifx\csname l@#1\endcsname\relax
\typeout{** WARNING: IEEEtran.bst: No hyphenation pattern has been}%
\typeout{** loaded for the language `#1'. Using the pattern for}%
\typeout{** the default language instead.}%
\else
\language=\csname l@#1\endcsname
\fi
#2}}
\providecommand{\BIBdecl}{\relax}
\BIBdecl

\bibitem{lagorce_hots_2017}
X.~Lagorce, G.~Orchard, F.~Galluppi, B.~E. Shi, and R.~B. Benosman, ``{HOTS}: {A} {Hierarchy} of {Event}-{Based} {Time}-{Surfaces} for {Pattern} {Recognition},'' \emph{IEEE Trans. Pattern Anal. Mach. Intell.}, vol.~39, no.~7, pp. 1346--1359, 2017.

\bibitem{rebecq_high_2019}
H.~Rebecq, R.~Ranftl, V.~Koltun, and D.~Scaramuzza, ``High speed and high dynamic range video with an event camera,'' \emph{{IEEE} Trans. Pattern Anal. Mach. Intell.}, vol.~43, no.~6, pp. 1964--1980, 2019.

\bibitem{arandjelovic_netvlad_2018}
R.~Arandjelovic, P.~Gronat, A.~Torii, T.~Pajdla, and J.~Sivic, ``{NetVLAD}: {CNN} {Architecture} for {Weakly} {Supervised} {Place} {Recognition},'' \emph{IEEE Trans. Pattern Anal. Mach. Intell.}, vol.~40, no.~6, pp. 1437--1451, 2018.

\bibitem{berton_rethinking_2022}
G.~Berton, C.~Masone, and B.~Caputo, ``\BIBforeignlanguage{en}{Rethinking {Visual} {Geo}-{Localization} for {Large}-{Scale} {Applications}},'' in \emph{\BIBforeignlanguage{en}{IEEE/CVF Conf. Comput. Vis. Pattern Recognit.}}, 2022, pp. 4878--4888.

\bibitem{ali-bey_mixvpr_2023}
A.~Ali-Bey, B.~Chaib-Draa, and P.~Giguere, ``{MixVPR}: {Feature} {Mixing} for {Visual} {Place} {Recognition},'' in \emph{IEEE/CVF Winter Conf. Appl. Comput. Vis.}, 2023, pp. 2997--3006.

\bibitem{berton_megaloc_2025}
G.~Berton and C.~Masone, ``Megaloc: One retrieval to place them all,'' in \emph{IEEE/CVF Conf. Comput. Vis. Pattern Recognit. Workshops}, 2025, pp. 2861--2867.

\bibitem{falanga_dynamic_2020}
D.~Falanga, K.~Kleber, and D.~Scaramuzza, ``\BIBforeignlanguage{en}{Dynamic obstacle avoidance for quadrotors with event cameras},'' \emph{\BIBforeignlanguage{en}{Sci. Robot.}}, vol.~5, no.~40, Mar. 2020.

\bibitem{hines_compact_2025}
A.~D. Hines, M.~Milford, and T.~Fischer, ``A compact neuromorphic system for ultra–energy-efficient, on-device robot localization,'' \emph{Sci. Robot.}, vol.~10, no. 103, p. eads3968, Jun. 2025.

\bibitem{gallego_event-based_2022}
G.~Gallego \emph{et~al.}, ``Event-{Based} {Vision}: {A} {Survey},'' \emph{IEEE Trans. Pattern Anal. Mach. Intell.}, vol.~44, no.~1, pp. 154--180, 2022.

\bibitem{lee_eventvlad_2021}
A.~J. Lee and A.~Kim, ``{EventVLAD}: {Visual} {Place} {Recognition} with {Reconstructed} {Edges} from {Event} {Cameras},'' in \emph{IEEE/RSJ Int. Conf. Intell. Robots Syst.}, Sep. 2021, pp. 2247--2252.

\bibitem{fischer_event-based_2020}
T.~Fischer and M.~Milford, ``Event-{Based} {Visual} {Place} {Recognition} {With} {Ensembles} of {Temporal} {Windows},'' \emph{IEEE Robot. Autom. Lett.}, vol.~5, no.~4, pp. 6924--6931, 2020.

\bibitem{lee_ev-reconnet_2023}
H.~Lee and H.~Hwang, ``Ev-{ReconNet}: {Visual} {Place} {Recognition} {Using} {Event} {Camera} {With} {Spiking} {Neural} {Networks},'' \emph{IEEE Sens. J.}, vol.~23, no.~17, pp. 20\,390--20\,399, 2023.

\bibitem{fischer_how_2022}
T.~Fischer and M.~Milford, ``How {Many} {Events} {Do} {You} {Need}? {Event}-{Based} {Visual} {Place} {Recognition} {Using} {Sparse} {But} {Varying} {Pixels},'' \emph{IEEE Robot. Autom. Lett.}, vol.~7, no.~4, pp. 12\,275--12\,282, 2022.

\bibitem{kong_event-vpr_2022}
D.~Kong \emph{et~al.}, ``Event-{VPR}: {End}-to-{End} {Weakly} {Supervised} {Deep} {Network} {Architecture} for {Visual} {Place} {Recognition} {Using} {Event}-{Based} {Vision} {Sensor},'' \emph{IEEE Trans. Instrum. Meas.}, vol.~71, pp. 1--18, 2022.

\bibitem{masone_survey_2021}
C.~Masone and B.~Caputo, ``A {Survey} on {Deep} {Visual} {Place} {Recognition},'' \emph{IEEE Access}, vol.~9, pp. 19\,516--19\,547, 2021.

\bibitem{lowry_visual_2016}
S.~Lowry \emph{et~al.}, ``Visual {Place} {Recognition}: {A} {Survey},'' \emph{IEEE Trans. Robot.}, vol.~32, no.~1, pp. 1--19, 2016.

\bibitem{zhang_visual_2021}
X.~Zhang, L.~Wang, and Y.~Su, ``\BIBforeignlanguage{en}{Visual place recognition: {A} survey from deep learning perspective},'' \emph{\BIBforeignlanguage{en}{Pattern Recognit.}}, vol. 113, p. 107760, 2021.

\bibitem{schubert_visual_2024}
S.~Schubert, P.~Neubert, S.~Garg, M.~Milford, and T.~Fischer, ``Visual {Place} {Recognition}: {A} {Tutorial},'' \emph{IEEE Robotics \& Automation Magazine}, vol.~31, no.~3, pp. 139--153, 2024.

\bibitem{garg_where_2021}
S.~Garg, T.~Fischer, and M.~Milford, ``\BIBforeignlanguage{en}{Where {Is} {Your} {Place}, {Visual} {Place} {Recognition}?}'' in \emph{\BIBforeignlanguage{en}{Int. Joint Conf. Artif. Intell.}}, vol.~5, 2021, pp. 4416--4425.

\bibitem{lengyel_zero-shot_2021}
A.~Lengyel, S.~Garg, M.~Milford, and J.~C. van Gemert, ``\BIBforeignlanguage{en}{Zero-{Shot} {Day}-{Night} {Domain} {Adaptation} {With} a {Physics} {Prior}},'' in \emph{\BIBforeignlanguage{en}{IEEE/CVF Int. Conf. Comput. Vis.}}, 2021, pp. 4399--4409.

\bibitem{milford_seqslam_2012}
M.~J. Milford and G.~F. Wyeth, ``{SeqSLAM}: {Visual} route-based navigation for sunny summer days and stormy winter nights,'' in \emph{IEEE Int. Conf. Robot. Autom.}, 2012, pp. 1643--1649.

\bibitem{garg_seqmatchnet_2022}
S.~Garg, M.~Vankadari, and M.~Milford, ``\BIBforeignlanguage{en}{{SeqMatchNet}: {Contrastive} {Learning} with {Sequence} {Matching} for {Place} {Recognition} \& {Relocalization}},'' in \emph{\BIBforeignlanguage{en}{Proc. Conf. Robot. Learn.}}, 2022, pp. 429--443.

\bibitem{carmichael_dataset_2025}
S.~Carmichael, A.~Buchan, M.~Ramanagopal, R.~Ravi, R.~Vasudevan, and K.~A. Skinner, ``\BIBforeignlanguage{en}{Dataset and {Benchmark}: {Novel} {Sensors} for {Autonomous} {Vehicle} {Perception}},'' \emph{\BIBforeignlanguage{en}{Int. J. Robot. Res.}}, vol.~44, no.~3, pp. 355--365, 2025.

\bibitem{tsintotas_revisiting_2022}
K.~A. Tsintotas, L.~Bampis, and A.~Gasteratos, ``\BIBforeignlanguage{en-US}{The {Revisiting} {Problem} in {Simultaneous} {Localization} and {Mapping}: {A} {Survey} on {Visual} {Loop} {Closure} {Detection}},'' \emph{\BIBforeignlanguage{en-US}{IEEE Trans. Intell. Transp. Syst.}}, vol.~23, no.~11, 2022.

\bibitem{zhang_lidar-based_2024}
Y.~Zhang, P.~Shi, and J.~Li, ``{LiDAR}-{Based} {Place} {Recognition} {For} {Autonomous} {Driving}: {A} {Survey},'' \emph{ACM Comput. Surv.}, vol.~57, no.~4, pp. 106:1--106:36, 2024.

\bibitem{venon_millimeter_2022}
A.~Venon, Y.~Dupuis, P.~Vasseur, and P.~Merriaux, ``Millimeter {Wave} {FMCW} {RADARs} for {Perception}, {Recognition} and {Localization} in {Automotive} {Applications}: {A} {Survey},'' \emph{IEEE Trans. Intell. Veh.}, vol.~7, no.~3, pp. 533--555, Sep. 2022.

\bibitem{melekhin_mssplace_2025}
A.~Melekhin, D.~A. Yudin, I.~Petryashin, and V.~Bezuglyj, ``{MSSPlace}: {Multi}-{Sensor} {Place} {Recognition} {With} {Visual} and {Text} {Semantics},'' \emph{IEEE Access}, vol.~13, pp. 177\,098--177\,110, 2025.

\bibitem{komorowski_minkloc_2021}
J.~Komorowski, M.~Wysoczańska, and T.~Trzcinski, ``{MinkLoc}++: {Lidar} and {Monocular} {Image} {Fusion} for {Place} {Recognition},'' in \emph{Int. Joint Conf. Neural Netw.}, 2021, pp. 1--8.

\bibitem{hou_fe-fusion-vpr_2023}
K.~Hou, D.~Kong, J.~Jiang, H.~Zhuang, X.~Huang, and Z.~Fang, ``{FE}-{Fusion}-{VPR}: {Attention}-{Based} {Multi}-{Scale} {Network} {Architecture} for {Visual} {Place} {Recognition} by {Fusing} {Frames} and {Events},'' \emph{IEEE Robot. Autom. Lett.}, vol.~8, no.~6, pp. 3526--3533, 2023.

\bibitem{wang_event_2025}
H.~Wang \emph{et~al.}, ``Event {Camera} {Meets} {Resource}-{Aware} {Mobile} {Computing}: {Abstraction}, {Algorithm}, {Acceleration}, {Application},'' 2025, arXiv:2503.22943 [cs].

\bibitem{chakravarthi_recent_2025}
B.~Chakravarthi, A.~A. Verma, K.~Daniilidis, C.~Fermuller, and Y.~Yang, ``Recent {Event} {Camera} {Innovations}: {A} {Survey},'' in \emph{Eur. Conf. Comput. Vis. {Workshops}}, 2025, pp. 342--376.

\bibitem{milford_towards_2015}
M.~Milford, H.~Kim, S.~Leutenegger, and A.~Davison, ``\BIBforeignlanguage{en}{Towards {Visual} {SLAM} with {Event}-based {Cameras}},'' \emph{\BIBforeignlanguage{en}{The problem of mobile sensors workshop in conjunction with RSS}}, 2015.

\bibitem{gehrig_end--end_2019}
D.~Gehrig, A.~Loquercio, K.~G. Derpanis, and D.~Scaramuzza, ``End-to-{End} {Learning} of {Representations} for {Asynchronous} {Event}-{Based} {Data},'' in \emph{IEEE/CVF Int. Conf. Comput. Vis.}, 2019, pp. 5633--5643.

\bibitem{nair_enhancing_2024}
G.~B. Nair, M.~Milford, and T.~Fischer, ``Enhancing {Visual} {Place} {Recognition} via {Fast} and {Slow} {Adaptive} {Biasing} in {Event} {Cameras},'' in \emph{IEEE/RSJ Int. Conf. Intell. Robots Syst.}, 2024, pp. 3356--3363.

\bibitem{hansen_neural_1990}
L.~Hansen and P.~Salamon, ``Neural network ensembles,'' \emph{IEEE Trans. Pattern Anal. Mach. Intell.}, vol.~12, no.~10, pp. 993--1001, 1990.

\bibitem{dietterich_ensemble_2000}
T.~G. Dietterich, ``\BIBforeignlanguage{en}{Ensemble {Methods} in {Machine} {Learning}},'' in \emph{\BIBforeignlanguage{en}{Multiple {Classifier} {Systems}}}.\hskip 1em plus 0.5em minus 0.4em\relax Springer, 2000, pp. 1--15.

\bibitem{hausler_hierarchical_2020}
S.~Hausler and M.~Milford, ``Hierarchical {Multi}-{Process} {Fusion} for {Visual} {Place} {Recognition},'' in \emph{IEEE Int. Conf. Robot. Autom.}, 2020, pp. 3327--3333.

\bibitem{hausler_patch-netvlad_2021}
S.~Hausler, S.~Garg, M.~Xu, M.~Milford, and T.~Fischer, ``\BIBforeignlanguage{en}{Patch-{NetVLAD}: {Multi}-{Scale} {Fusion} of {Locally}-{Global} {Descriptors} for {Place} {Recognition}},'' in \emph{\BIBforeignlanguage{en}{IEEE/CVF Conf. Comput. Vis. Pattern Recognit.}}, 2021, pp. 14\,141--14\,152.

\bibitem{hussaini_ensembles_2023}
S.~Hussaini, M.~Milford, and T.~Fischer, ``Ensembles of {Compact}, {Region}-specific \& {Regularized} {Spiking} {Neural} {Networks} for {Scalable} {Place} {Recognition},'' in \emph{IEEE Int. Conf. Robot. Autom.}, 2023, pp. 4200--4207.

\bibitem{hu_ddd20_2020}
Y.~Hu, J.~Binas, D.~Neil, S.-C. Liu, and T.~Delbruck, ``{DDD20} end-to-end event camera driving dataset: Fusing frames and events with deep learning for improved steering prediction,'' in \emph{IEEE Int. Conf. Intell. Transp. Syst.}, 2020.

\bibitem{pan_nyc-event-vpr_2024}
T.~Pan, J.~He, C.~Chen, Y.~Li, and C.~Feng, ``Nyc-event-vpr: A large-scale high-resolution event-based visual place recognition dataset in dense urban environments,'' in \emph{IEEE Int. Conf. Robot. Autom.}, 2025, pp. 4657--4664.

\bibitem{berton_eigenplaces_2023}
G.~Berton, G.~Trivigno, B.~Caputo, and C.~Masone, ``\BIBforeignlanguage{en}{{EigenPlaces}: {Training} {Viewpoint} {Robust} {Models} for {Visual} {Place} {Recognition}},'' in \emph{\BIBforeignlanguage{en}{IEEE/CVF Int. Conf. Comput. Vis.}}, 2023, pp. 11\,080--11\,090.

\end{thebibliography}
\end{document}